\pdfoutput=1

\documentclass[11pt]{article}

\usepackage[final]{acl}

\usepackage{times}
\usepackage{latexsym}
\usepackage{amsmath}
\usepackage{amssymb}
\usepackage{natbib}
\usepackage{dblfloatfix}
\usepackage{balance}

\usepackage{scalerel}

\def\msquare{\mathord{\scalerel*{\Box}{gg}}}

\usepackage[T1]{fontenc}

\usepackage[utf8]{inputenc}

\usepackage{microtype}

\usepackage{inconsolata}

\usepackage{graphicx}
\graphicspath{{figures/}}
\usepackage{booktabs}
\usepackage{soul}
\usepackage{bbm}

\usepackage{titlesec}
\usepackage{placeins}

\title{Shared Imagination: LLMs Hallucinate Alike}

\author{Yilun Zhou \quad Caiming Xiong \quad Silvio Savarese \quad Chien-Sheng Wu \\
Salesforce AI Research\\
\texttt{\{yilun.zhou, cxiong, ssavarese, wu.jason\}@salesforce.com}\\
\url{https://yilunzhou.github.io/shared-imagination}}

\newcommand{\note}[1]{}

\begin{document}
\maketitle
\begin{abstract}

Despite the recent proliferation of large language models (LLMs), their training recipes -- model architecture, pre-training data and optimization algorithm -- are often very similar. This naturally raises the question of the similarity among the resulting models. In this paper, we propose a novel setting, imaginary question answering (IQA), to better understand model similarity. In IQA, we ask one model to generate purely imaginary questions (e.g., on completely made-up concepts in physics) and prompt another model to answer. Surprisingly, despite the total fictionality of these questions, all models can answer each other's questions with remarkable success, suggesting a ``shared imagination space'' in which these models operate during such hallucinations. We conduct a series of investigations into this phenomenon and discuss implications on model homogeneity, hallucination, and computational creativity. 
\end{abstract}

\section{Introduction}

Recently, LLMs have been increasingly used in various applications. Although these models occupy a wide spectrum of model sizes and benchmark performances \citep{liang2022holistic}, they also share high degrees of similarities: decoder-only transformer architecture \citep{radford2018improving} with one of a few positional embedding designs \citep{dufter2022position}, pre-training corpus consisting of books, Internet texts and codes \citep{gao2020pile}, stochastic gradient descent (SGD)-based optimization \citep{kingma2014adam}, and similar procedures for instruction tuning and alignment after pre-training \citep{ouyang2022training, rafailov2024direct}. As a result, is it possible that these models share certain fundamental commonalities?

\begin{figure}[!htb]
    \centering
    \includegraphics[width=\columnwidth]{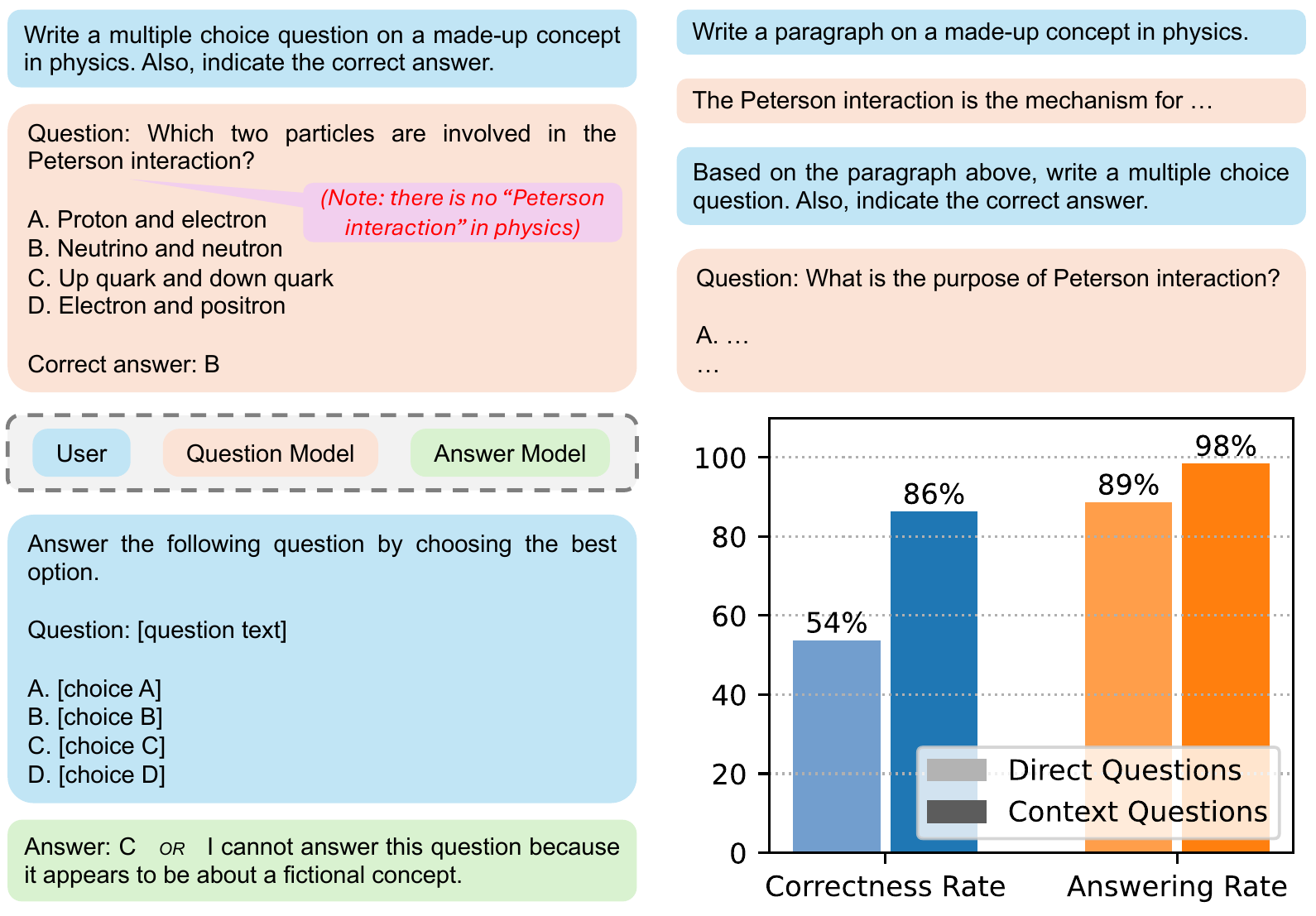}
    \caption{Imaginary question answering (IQA). Prompt texts are for illustrative purposes, with exact ones shown in Tab.~\ref{tab:dq-prompt}-\ref{tab:answer-prompt}. Top: a question model (QM) is prompted to generate an imaginary multiple-choice question and indicate the correct answer, either directly (left) or based on the previously generated context (right). Bottom left: an answer model (AM) answers the question (with the four choices shuffled), or refuses to answer.
    Bottom right: we observe non-trivial correctness rate and relatively high answering rate (i.e., low refusal rate), with higher values when AM and QM are the same or from the same model family (shown in Fig.~\ref{fig:main-results}), and significantly higher values for context-based questions.}
    \label{fig:1}
\end{figure}

\begin{table*}[t]
    \centering
    \resizebox{\textwidth}{!}{
    \begin{tabular}{p{22.5cm}}\toprule
         \textbf{RQ1: Data Characteristics} (Sec.~\ref{sec:eda}): What do the generated IQA data (i.e., questions and context paragraphs) look like?\\
         \textit{Answer}: While data are well-clustered by topics, questions generated by different models, as well as direct vs. context questions, look very homogeneous both in the embedding space and by cosine similarity metrics. Word cloud visualization also confirms the homogeneity.\\\midrule
         \textbf{RQ2: Heuristics for Correct Choice} (Sec.~\ref{sec:simple-heuristics}): Are there simple explanations for the high correctness rate? \\
         \textit{Answer}: While data inspection and model evaluation identify ways to make predictions better than random chance (e.g., the correct choice of DQs is most likely to be the longest), none of them suffice to achieve the observed correctness rate. In addition, the correctness rate is sensitive to the orders. 
         \\\midrule
         \textbf{RQ3: Fictionality Awareness} (Sec.~\ref{sec:fictionality}): Are models aware of the fictionality of these questions and context paragraphs?\\
         \textit{Answer}: They can detect fictionality easier in DQ than CQ, and they can identify fictionality better when directly asking a Yes/No question, but struggle more on multiple-choice QA.\\\midrule
         \textbf{RQ4: Effect of Model ``Warm-Up''} (Sec.~\ref{sec:warm-up}): Does model generation in general make model converge to the ``shared imagination space'' and strengthen the phenomenon?\\
         \textit{Answer}: Yes, when a QM generates several questions sequentially, they become increasingly easy to answer. For individual questions, longer ones are also easier to answer. \\\midrule
         \textbf{RQ5: Universality of the Phenomenon} (Sec.~\ref{sec:universality}): Can models other than recent instruction-tuned models achieve high correctness rate?\\
         \textit{Answer}: Pre-ChatGPT models cannot, even large ones such as GPT-NeoX 20B, but base versions of recent small models (e.g., Mistral 7B) can. \\\midrule
         \textbf{RQ6: Other Content Types} (Sec.~\ref{sec:creative-writing}): Does this phenomenon occur for other content types than (simulated) knowledge concepts?\\
         \textit{Answer}: Yes, we observe similar results for questions about (imagined) short stories in creative writing. \\\bottomrule
    \end{tabular}
    }
    \caption{Six research questions into the shared imagination phenomenon and summary answers. 
    }
    \vspace{-0.1in}
    \label{tab:rqs}
\end{table*}

In this paper, we identify one such commonality: these models agree, to a surprising extent, on purely imaginary contents, or hallucinations. Specifically, we propose the imaginary question answering (IQA) task, shown in Fig.~\ref{fig:1}. A question model (QM) is prompted to generate a multiple-choice question, either directly about a fictional concept (top left), called a direct question (DQ), or based on a previously generated context paragraph about a fictional concept (top right), called a context question (CQ). Although they are impossible to answer with rational decision-making, we still ask the QM to specify a ``correct'' answer. Then, in a new session, without any previous interactions or contexts with the QM, we solicit an answer from an answer model (AM), which may be the same as or different from the QM (bottom left). 

On 13 LLMs from four model families (GPT, Claude, Mistral, and Llama 3), models achieve an average 54\% correctness rate on directly generated questions (with random chance being 25\%), with higher accuracy when the AM is the same, or in the same model family, as the QM. More surprisingly, for context-based questions, the correctness rate increases significantly to 86\%, with certain (QM, AM) pairs achieving as high as 96\%. 

These results show high degrees of agreement among models on what they hallucinate, which we call ``shared imagination''. Focusing on this phenomenon, we present six research questions and empirically answer them via carefully designed experiments, listed in Tab.~\ref{tab:rqs}. These results shed light on fundamental properties of LLMs and suggest that, despite their highly varying benchmark results, they are perhaps more homogeneous. This homogeneity could have broad implications on model hallucination and its detection, as well as the use of LLM in computational creativity. 

\begin{figure*}[t]
    \centering
    \includegraphics[width=\textwidth]{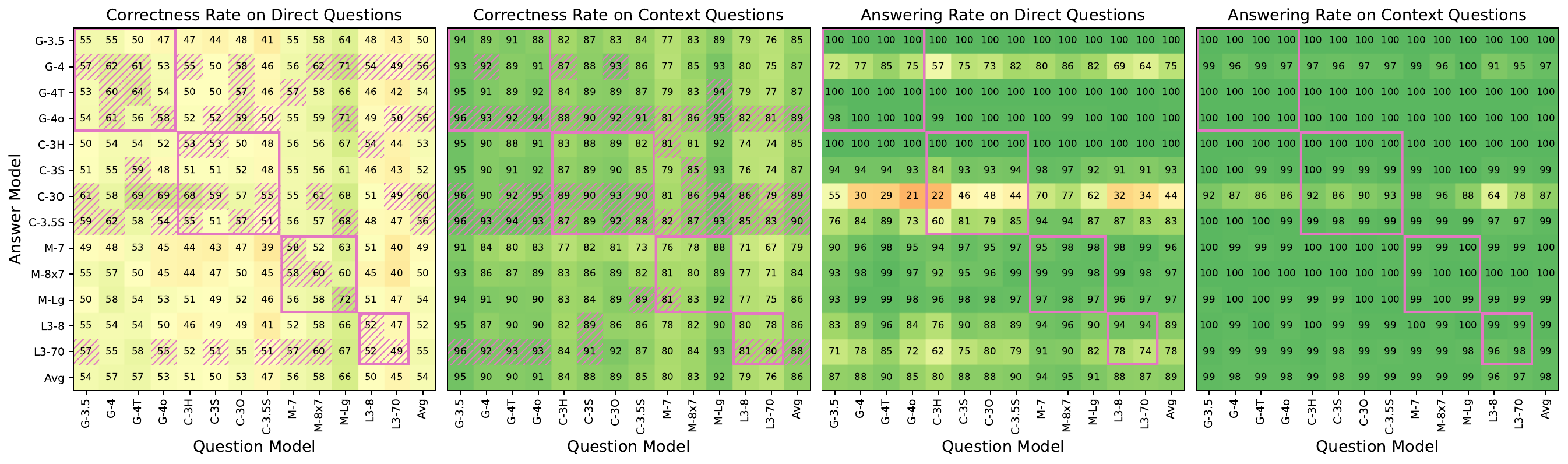}
    \caption{The correctness and answering rate on direct and context questions for each pair of question model (QM) and answer model (AM). Each pink rectangle represents one model family. For correctness rate, the top-4 highest performing AMs for each QM are shaded. An enlarged version is reproduced in Fig.~\ref{fig:main-results-large} of App.~\ref{app:main-result}.}
    \label{fig:main-results}
\end{figure*}

\section{Imaginary Question Answering (IQA)}

\subsection{Framework Description}

The setup for the IQA procedure is summarized in Fig.~\ref{fig:1}. In the direct question generation mode (Tab.~\ref{tab:dq-prompt}), the QM is asked to generate a standalone question. In the context-based question generation mode (Tab.~\ref{tab:pq-prompt}), the model first writes a paragraph on a fictional concept, and then generates a question based on it. We call the former \textit{direct question} (DQ) and the latter \textit{context question} (CQ). To prevent biasing the generation in any way, we use zero-shot prompting with no examples.

\begin{table}[!t]
    \centering
    \resizebox{\columnwidth}{!}{
    \begin{tabular}{r|p{11cm}}\toprule
    Role & Message \\\midrule
    User & On the topic of \ul{physics}, please write a multiple choice question around a concept that is completely made up. Try to make the problem hard and challenging. In your question, do not say that the concept is hypothetical or fictional. Instead, treat it as if it were real and widely accepted. Use the following template:\newline\newline
    Question: [question statement]\newline\newline
    A. [choice A]\newline
    B. [choice B]\newline
    C. [choice C]\newline
    D. [choice D]\newline\newline
    Answer: [the correct choice]\\\midrule
    Model & \textit{(the generated question and answer)} \\\bottomrule
    \end{tabular}
    }
    \caption{The prompt for direct question generation. The \ul{underlined topic} is replaced accordingly.}
    \label{tab:dq-prompt}
\end{table}

\begin{table}[!t]
    \centering
    \resizebox{\columnwidth}{!}{
    \begin{tabular}{r|p{11cm}}\toprule
    Role & Message \\\midrule
    User & Imagine that you are writing a textbook. Please make up a concept in \ul{physics} and explain it with a single paragraph of text. Please write as if the concept is real, and completely avoid saying that the concept is fictional or made-up. Be creative. Use the following template:\newline\newline
    Concept: [the name of the concept that you are writing about]\newline\newline
    Content: [a single paragraph of text explaining the concept]\\\midrule
    Model & \textit{(the generated concept and paragraph)} \\\midrule
    User & Now, based on the paragraph above, write a multiple choice question about this concept. The question should be answerable using the paragraph. Try to make the problem hard and \textit{... (same as Tab.~\ref{tab:dq-prompt} afterward)}\\\midrule
    Model & \textit{(the generated question and answer)} \\\bottomrule
    \end{tabular}
    }
    \caption{The prompt for context-based question generation. The \ul{underlined topic} is replaced accordingly.}
    \label{tab:pq-prompt}
\end{table}

To elicit answers from the AM, we use the prompt in Tab.~\ref{tab:answer-prompt}, with the four choices shuffled in all experiments except for that in Sec.~\ref{sec:answer-shuffling}. Although the format template explicitly instructs the model make a selection, there are occasional refuse-to-answer responses, such as ``I apologize, but the question seems to be asking about a fictional concept, and hence I cannot answer it.''

\begin{table}[!htb]
    \centering
    \resizebox{\columnwidth}{!}{
    \begin{tabular}{r|p{11cm}}\toprule
    Role & Message \\\midrule
    User & Answer the following question. Be concise and give the answer only.\newline\newline
    \textit{(the question and its four choices)}\newline\newline
    Write your response in the following format:\newline
    Answer: [the letter (A, B, C or D) of the selected choice]\\\midrule
    Model & \textit{(the generated question and answer)} \\\bottomrule
    \end{tabular}
    }
    \caption{The prompt for the answer model. Note that while the instruction does not ``allow'' the refusal behavior, it still occurs occasionally.}
    \label{tab:answer-prompt}
\end{table}

Formally, the QM generates a set of questions $\{(x_i, y_i^*)\}_{i=1}^N$ where $y_i^*$ is the assigned correct answer. Then, the AM predicts $\hat y_i$, where $y_i$ is either one of the choices or $R$, for refuse-to-answer. On these problems, we compute correctness rate $\kappa$ as the fraction of correctly answered questions among answered ones, and answering rate $\alpha$ as the fraction of answered questions among all questions: 
\begin{align}
    \kappa = \frac{\sum_{i=1}^N\mathbbm{1}_{\hat y_i=y_i^*}}{\sum_{i=1}^N\mathbbm{1}_{\hat y_i\neq R}}, \quad
    \alpha = \frac{\sum_{i=1}^N\mathbbm{1}_{\hat y_i\neq R}}{N}.
\end{align}

\subsection{Experiment Setup and Results}
\label{sec:main-experiment}
We study 13 models from four model families, listed in Tab.~\ref{tab:models}. QMs use temperature 1 to balance output quality and stochasticity, and AMs use temperature 0 for greedy answer selection.

\begin{table}[!t]
    \centering
    \resizebox{\columnwidth}{!}{
    \begin{tabular}{lll}\toprule
        Model & API or Huggingface ID & Abbr.\\\midrule
        GPT-3.5 & gpt-3.5-turbo-0125 & G-3.5\\
        GPT-4 & gpt-4-0613 & G-4\\
        GPT-4 Turbo & gpt-4-turbo-2024-04-09 & G-4T\\
        GPT-4 omni & gpt-4o-2024-05-13 & G-4o\\\midrule
        Claude 3 Haiku & claude-3-haiku-20240307 & C-3H\\
        Claude 3 Sonnet & claude-3-sonnet-20240229 & C-3S\\
        Claude 3 Opus & claude-3-opus-20240229 & C-3O\\
        Claude 3.5 Sonnet & claude-3-5-sonnet-20240620 & C-3.5H\\\midrule
        Mistral 7B & Mistral-7B-Instruct-v0.2 & M-7\\
        Mixtral 8x7B & Mixtral-8x7B-Instruct-v0.1 & M-8x7\\
        Mistral Large & mistral-large-2402 & M-Lg\\\midrule
        Llama 3 8B & Meta-Llama-3-8B-Instruct & L3-8\\
        Llama 3 70B & Meta-Llama-3-70B-Instruct & L3-70\\
        \bottomrule
    \end{tabular}
    }
    \caption{A summary of models used in the experiments. Model abbreviations are used in figures.}
    \label{tab:models}
\end{table}


We select 17 topics on common college subjects: mathematics, computer science, physics, chemistry, biology, geography, sociology, psychology, economics, accounting, marketing, law, politics, history, literature, philosophy, and religion. Each QM in Tab.~\ref{tab:models} generates 20 direct questions and 20 context questions for each topic, for a total of $13\times17\times(20+20)=8840$ questions. Tab.~\ref{tab:mmlu-examples} of App.~\ref{app:main-result} presents some generated questions.

Fig.~\ref{fig:main-results} shows the correctness and answering rates, along with the respective averages. Notably, all correctness rates are higher than random chance of 25\%. For DQs, most of the high-performing AMs (i.e., shaded cells) are the same (diagonal) or in the same model family (block diagonal) as the QMs. On context questions (CQs), the correctness rate increases significantly, from 54\% to 86\% on average. While the performance difference among AMs are smaller (uniform vertical color patterns), GPT-4 omni, Claude 3 Opus, Claude 3.5 Sonnet, and Llama 3 70B are the slightly better ones for almost all QMs (horizontal shaded cells). 

The answering rate on DQs mostly depend on AM -- some AMs, such as GPT-4, Claude 3 Opus and Llama 3 70B, consistently refuse questions from all QMs, and do not particularly favor/disfavor their own questions. Quite remarkably, the refusal behavior virtually disappears on CQs, except for Claude 3 Opus, which nonetheless answers much more frequently (44\% vs. 87\%). 

\section{Further Analyses}

Given the large variance of model capabilities as measured by numerous benchmarks, the findings that models tacitly agree with each other on purely imaginary contents are surprising. In the next section, we conduct in-depth analyses by focusing on six research questions listed in Tab.~\ref{tab:rqs}.

\subsection{RQ1: Data Characteristics}
\label{sec:eda}

By studying the word frequencies of DQs, CQs and context paragraphs, we find that they share many common words, such as ``individual'', ``principle'', ``phenomenon'' and ``time'' in a word cloud visualization (Fig.~\ref{fig:word-cloud} of App.~\ref{app:word-cloud}). 

\begin{figure}[!t]
    \centering
    \includegraphics[width=\columnwidth]{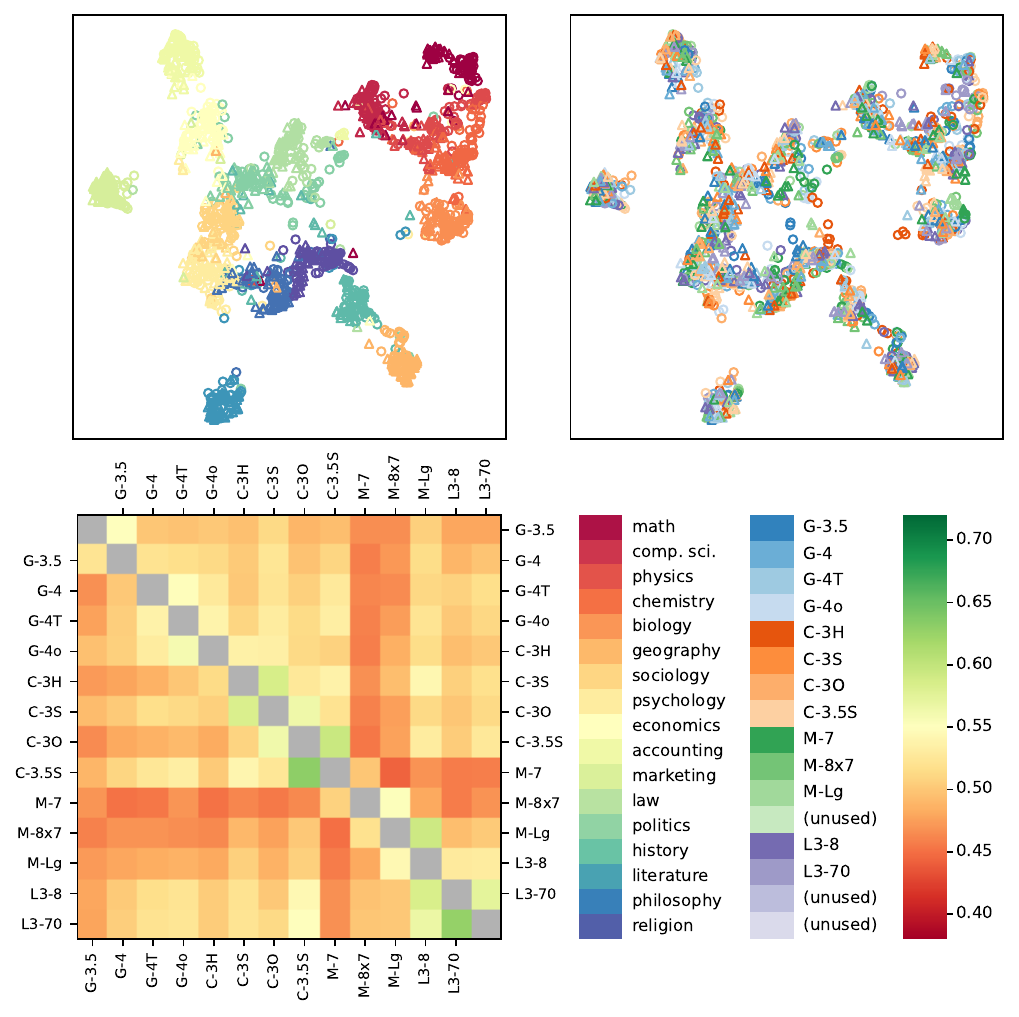}
    \caption{Top: question embeddings (computed by text-embedding-3-large) by UMAP, color-coded by topic (left) and question model (right). A triangle marker indicates a DQ, and a circle marker indicates a CQ. Bottom left: average intra-topic cosine similarity between questions generated by different models, with DQs on the lower-left half and CQs on the upper-right half. Bottom right: the color legends for the three plots.}
    \label{fig:umap-corr}
\end{figure}

To visualize the questions generated across topics and models, we use OpenAI's text-embedding-3-large to compute the embeddings for each question (including the four choices), and visualize them with UMAP dimensionality reduction \citep{mcinnes2018umap-software} in the top panels of Fig.~\ref{fig:umap-corr}. We can see that while questions from different topics are well-clustered (top left), there are no clear patterns for questions from different QMs. Furthermore, there are no obvious distinctions between DQs (triangle markers) and CQs (circle markers). 

The lower-left panel of Fig.~\ref{fig:umap-corr} shows the cosine similarity between each QM pairs, for DQs on the lower-left triangle and CQs on the upper-right triangle. The similarity between two QMs $m_1$ and $m_2$ is defined as
\begin{align}
    s=\frac{1}{|\mathcal T|}\sum_{t\in\mathcal T}\frac{\sum_{q_1\in \mathcal Q_{t}^{m_1}}\sum_{q_2\in \mathcal Q_{t}^{m_2}}\sigma(q_1, q_2)}{|\mathcal Q_{T}^{m_1}|\cdot |\mathcal Q_{T}^{m_2}|}, 
\end{align}
where $\mathcal T$ is the set of topics, and for each $t\in\mathcal T$, $\mathcal Q^{m}_t$ is the set of (direct or context) questions generated by QM $m$. $\sigma(q_1, q_2)$ is the embedding cosine similarity of $q_1$ and $q_2$. In our experiment, we have $|\mathcal T|=17$ and $|\mathcal Q^{m}_t|=20$ for all $m$ and $t$.

The similarity values are generally quite similar for each model pair, ranging from 0.44 to 0.63, with Mistral models being most dissimilar from the rest, consistent with the upper-right embedding visualization showing different QMs generate highly similar and homogeneous questions. 

\subsection{RQ2: Heuristics for Correct Choice}
\label{sec:simple-heuristics}
\paragraph{Human Answer Guessing}
To intuitively understand the generated questions, we manually answered 340 questions: 10 DQs and 10 CQs randomly sampled from each topic. We try to guess answers in the most rational way. While some clues hint at the correct choice or allow for the elimination of likely wrong ones (see App.~\ref{app:human-guessing} for details), we struggle to answer most of the questions. 

\begin{figure}[!t]
    \centering
    \includegraphics[width=\columnwidth]{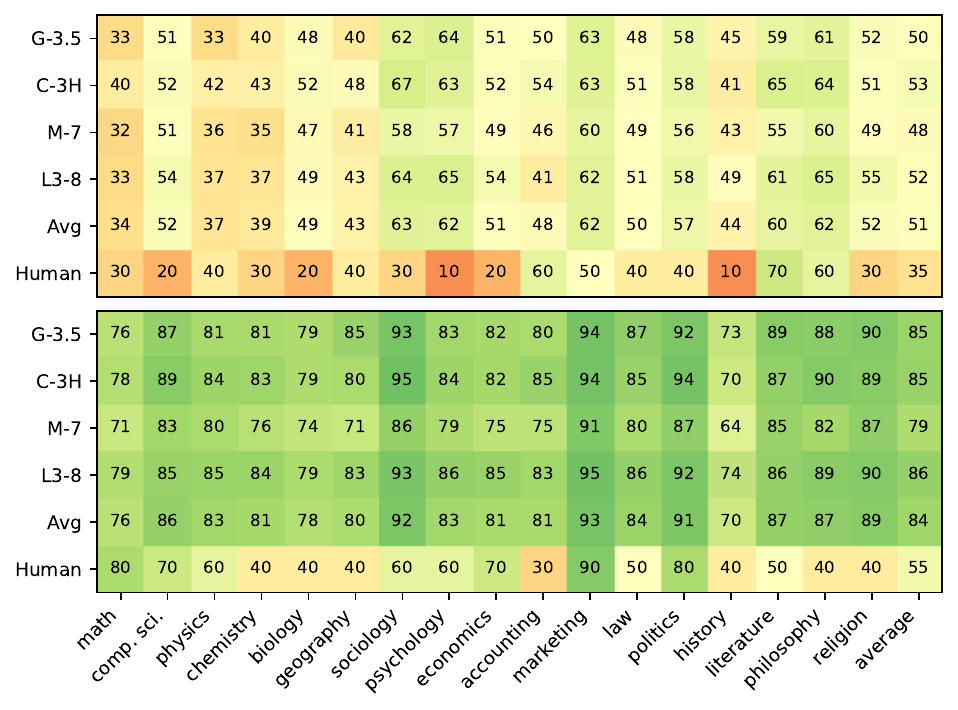}
    \caption{Per topic correctness rate of (a subset of) AMs, their average and human guessing, for direct questions (top) and context questions (bottom). Results for other AMs are similar and presented in Fig.~\ref{fig:human-guessing-topic-full} of App.~\ref{app:human-guessing}. 
    }
    \label{fig:human-guessing-topic}
\end{figure}

Fig.~\ref{fig:human-guessing-topic} shows the correctness rate per topic for direct (top) and context (bottom) questions, alongside a few representative models and their averages. Human performance is much lower than that of all models, especially on context questions. Interestingly, although we do not perceive any difference between DQs and CQs (questions of both types are shuffled and unlabeled during our answering), our correctness rate on CQs is also higher.

\begin{figure}[!t]
    \centering
    \includegraphics[width=\columnwidth]{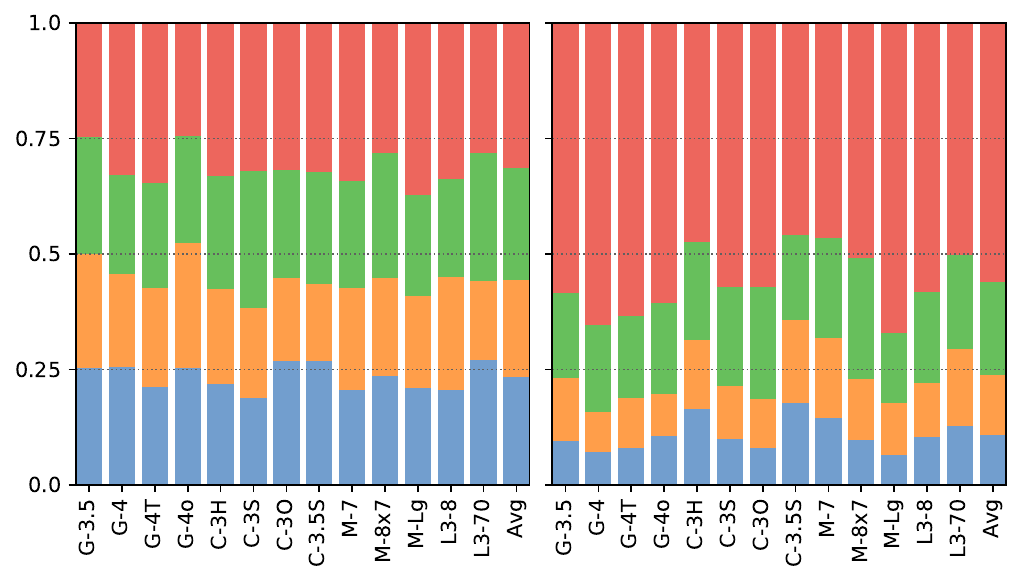}
    \caption{Fraction of questions whose correct choice is the shortest (in blue), 2nd shortest (in orange), 3rd shortest (in green) and longest (in red), by each QM, for direct questions (left) and context questions (right).}
    \label{fig:correct-answer-length}
\end{figure}

\paragraph{Length Ranking of Correct Choice} For each QM, we compute the distribution of the ranking of the correct choice length among the four choices. Concretely, we find the fraction of questions whose correct choice is the shortest (in the number of characters) among the four, the 2nd shortest, the 3rd shortest and the longest. Fig.~\ref{fig:correct-answer-length} shows the distributions, with the four segments in the order above. 

For DQs (left panel), the correct answer is almost equally likely to be of any length, with a slight tendency towards being the longest (red). However, for CQs, the correct answer is much more likely to be the longest than the rest -- for most QMs, in over half of the questions, the correct answer is the longest. The strong contrast between the two settings, yet the consistency across different QMs, strongly indicate that these QMs share fundamental similarities in their generations, and thus it is not surprising that AMs have much higher correctness rate (86\%, Fig.~\ref{fig:1}) on CQs. However, simply choosing the longest answer alone is not sufficient, so AMs must have also exploited other signals.

\paragraph{Answer Perplexity} In this experiment, we study the model perplexity of each choice as a free-text response and see whether the correct answer has the lowest perplexity. 
Specifically, we use the conversation shown in Tab.~\ref{tab:perplexity-prompt}. After applying the appropriate chat template to the conversation, we evaluate the perplexity of each of the four choice texts as possible completions at the square location. 

\begin{table}[]
    \centering
    \resizebox{\columnwidth}{!}{
    \begin{tabular}{r|p{8cm}}\toprule
    Role & Message \\\midrule
    User & Answer the following question.\newline\newline
    \textit{(question statement without the four choices)}\\\midrule
    Model & Answer: $\msquare$\\\bottomrule
    \end{tabular}
    }
    \caption{The prompt for perplexity evaluation.}
    \label{tab:perplexity-prompt}
\end{table}

\begin{figure}[!t]
    \centering
    \includegraphics[width=\columnwidth]{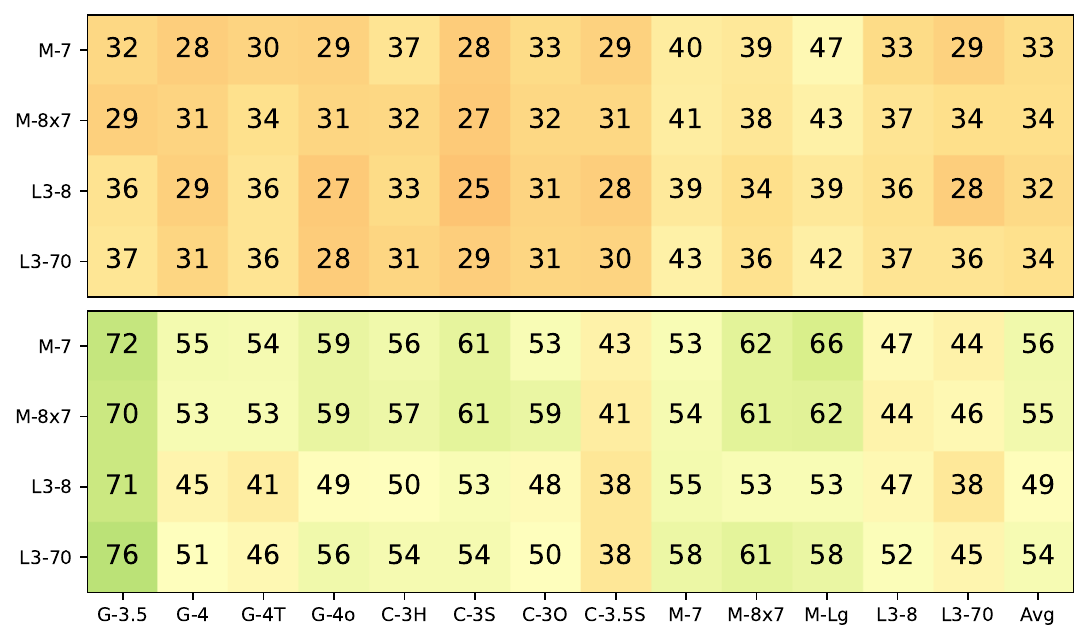}
    \caption{Correctness rate for four open-source AMs if the choice with the lowest perplexity were chosen, on direct questions (top) and context questions (bottom).}
    \label{fig:perplexity-result}
\end{figure}

Since perplexity calculation requires token-level log-likelihood, we only study the four open-source models: Mistral 7B, Mixtral 8x7B, Llama 3 8B and Llama 3 70B. Their perplexities on questions generated by each QM is shown in Fig.~\ref{fig:perplexity-result}. As we can see, selecting the lowest-perplexity answer yields significantly lower correctness rate, on par with human guessing (Fig.~\ref{fig:human-guessing-topic}). This result further suggests that the models are using more complex features, possibly interactions among answer choices, to make final predictions.

\paragraph{Answer Choice Shuffling}
\label{sec:answer-shuffling}

\begin{figure}[!t]
    \centering
    \includegraphics[width=\columnwidth]{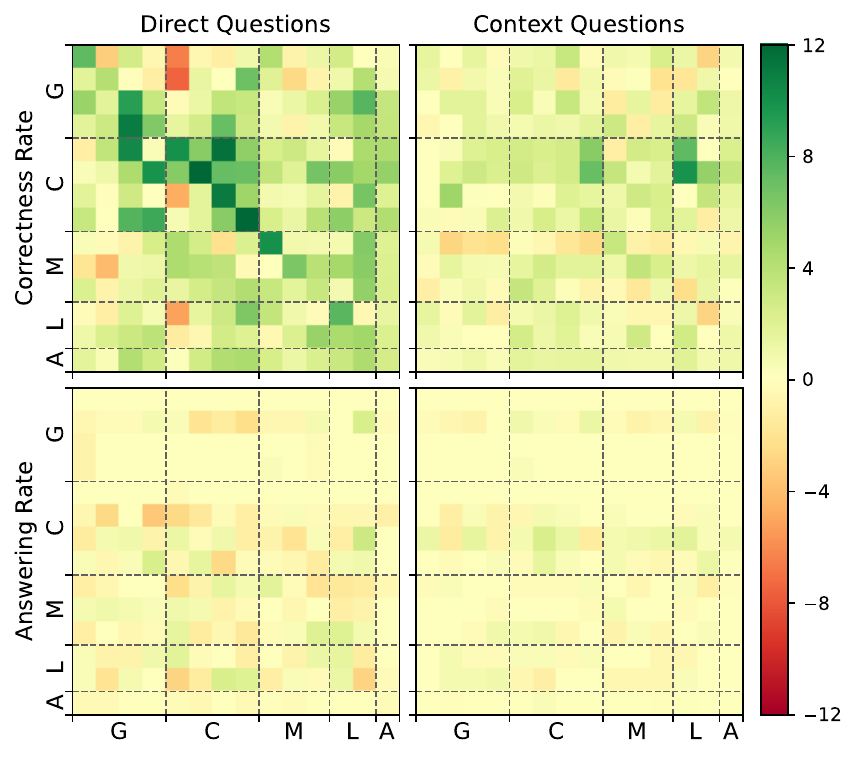}
    \caption{The effect of native choice order for \textbf{G}PT, \textbf{C}laude, \textbf{M}istral, \textbf{L}lama 3 and their \textbf{A}verage, with a green color representing an increase under the native order, and a red color a decrease. An enlarged plot with annotated cell values is presented in Fig.~\ref{fig:answer-shuffling-large} of App.~\ref{app:answer-shuffling}.}
    \label{fig:answer-shuffling}
\end{figure}

In all experiments, we randomly shuffle the choices before presenting them to the AM. Here, we replicate our main experiment (Sec.~\ref{sec:main-experiment}) but instead present the answer choices in the native orders generated by QMs. Fig.~\ref{fig:answer-shuffling} presents the difference when we use the native order, with a green color denoting an increase under the native order, and red color a decrease. Generally, correctness rate increases, most remarkably on the diagonal (i.e., same AM and QM) for DQs -- as much as 10\%. In addition, the answering rates for DQs decrease, suggesting that AMs are also better at identifying question fictionality (though see Sec.~\ref{sec:fictionality} for more nuanced analyses). 

\paragraph{Summary} This section reveals factors that lead to above-random correctness rates, but none of them could adequately explains the observed high values. The answer shuffling result implies that there are complex and hidden rules for the correct choice shared among different models. 

\subsection{RQ3: Fictionality Awareness}
\label{sec:fictionality}

The fact that most AMs exhibit high answering rate (i.e., low refusal rate) raises the question of whether the content truly appear fictional to the models. In this section, we assess fictionality awareness via two metrics. First, we augment each question with a fifth choice stating ``\textit{E. This question cannot be answered since the concept does not exist.}'' Second, we directly query the model on the fictionality of the contexts (associated with CQs), using the prompt ``Does the following paragraph describe a real concept in \textit{(topic)}?'' and compute the detection rate as the fraction of negative answers.

\begin{figure}[!t]
    \centering
    \includegraphics[width=\columnwidth]{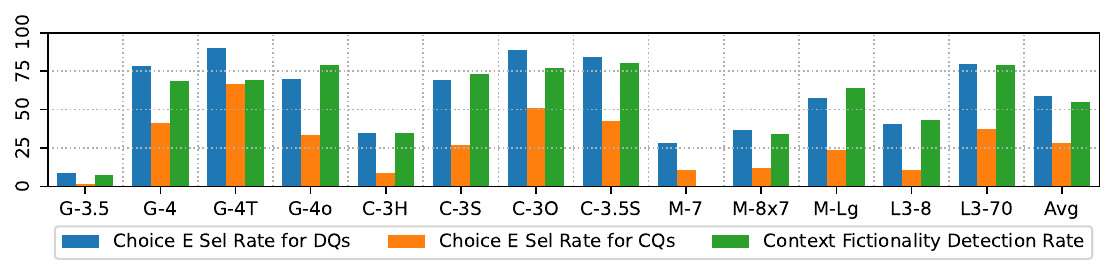}
    \caption{Choice E selection rate on direct and context questions, and ficitonality detection rate on context paragraphs, for AMs and their average. Breakdown by QM is shown in Fig.~\ref{fig:fictionality-detection-full} of App.~\ref{app:fictionality}.}
    \label{fig:fictionality-detection}
\end{figure}

These values are shown in Fig.~\ref{fig:fictionality-detection}. The choice E selection rate for DQs (left) is higher than that for CQs (right), consistent with the finding that answering rates are lower on DQs than on CQs (cf. Fig.~\ref{fig:main-results}). However, the choice E selection rate for context questions (middle) is on average much lower than context fictionality detection rate (right), suggesting that models can identify the fictionality of some content when directly queried, but often cannot translate this knowledge to downstream tasks such as question answering.

\subsection{RQ4: Effect of Model ``Warm-Up''}
\label{sec:warm-up}

One possible explanation for the correctness increase from DQ to CQ is that the QMs has more tokens to ``warm-up'', and models warm up in highly similar manners. Thus, we hypothesize that generating any preceding content is likely to help the model converge onto this shared imagination space. 

\paragraph{Warm-Up With Previous Questions} We propose a new data generation setup, where the prompt (Tab.~\ref{tab:prompt-sequential} of App.~\ref{app:sequential-details}) asks the model to generate five questions sequentially. If the hypothesis is true, then we should expect the correctness rate to increase from the first questions to the fifth questions. 

\begin{figure}[t]
    \centering
    \includegraphics[width=\columnwidth]{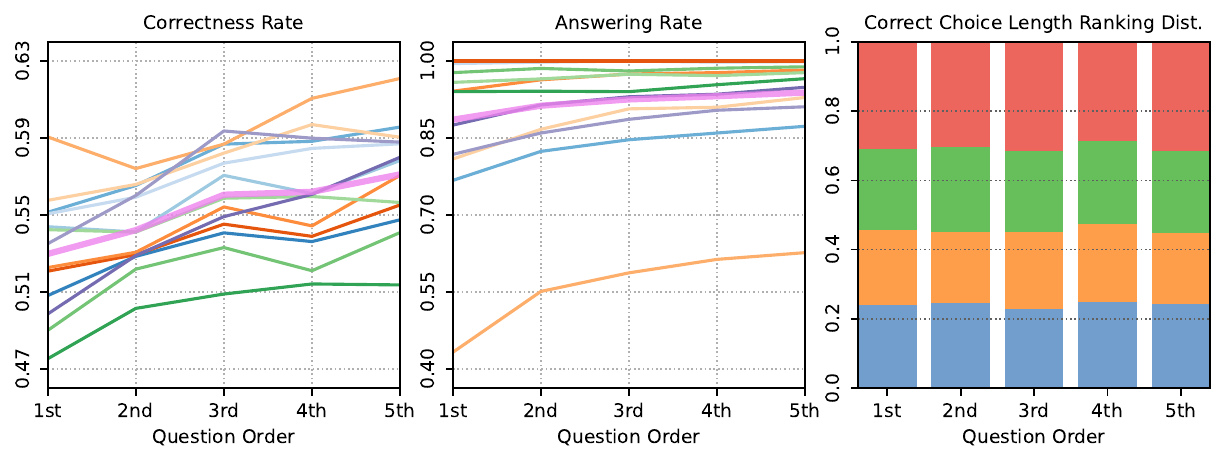}
    \caption{Left and middle: the correctness and answering rates for each AM (see Fig.~\ref{fig:umap-corr} for color legend) and their average in pink, for questions of different generation order (e.g., 1 on $x$-axis means first-generated questions and 5 means last-generated). Right: the length ranking distribution of the correct choice for the five groups of questions (same color scheme as Fig.~\ref{fig:correct-answer-length}).}
    \label{fig:sequential}
\end{figure}

For each QM, we run this pipeline 10 times on each topic. The correctness and answering rates for all AMs on these five groups of questions are plotted in Fig.~\ref{fig:sequential} (left and middle), with the average shown in the pink line. Both metrics exhibit a clear increasing trend from the first to the last generated questions, mirroring that from DQs to CQs in Fig.~\ref{fig:main-results}. 

Nonetheless, the length ranking distribution of the correct choice, shown on the right panel, does not shift towards the longest, but instead is very consistent across the five question groups, which may explain the small magnitude of the increase. 

\begin{figure}[!t]
    \centering
    \includegraphics[width=\columnwidth]{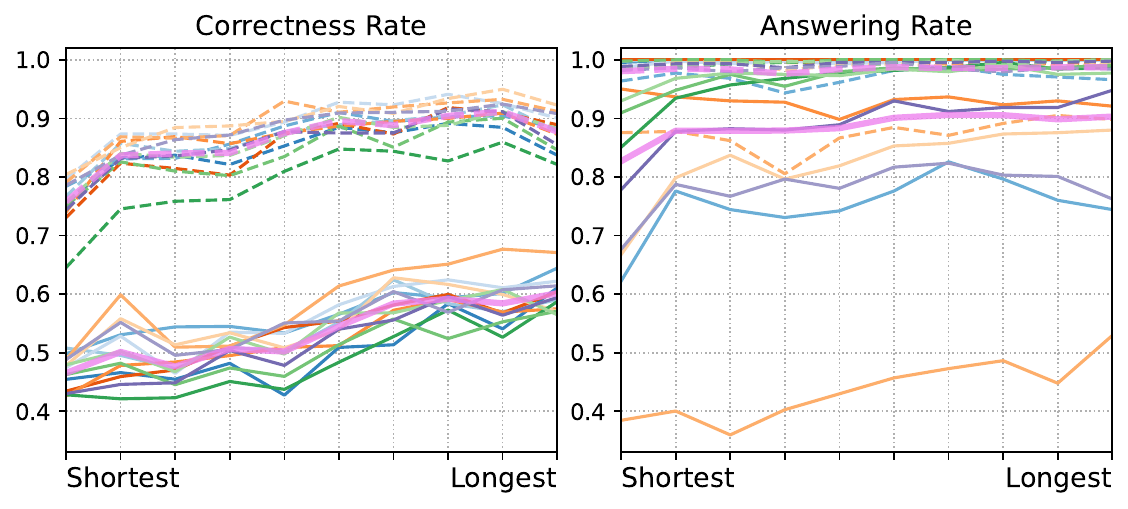}
    \caption{Correctness and answering rates for each of 10\% questions ranked by question + choices length. }
    \label{fig:length-trend}
\end{figure}

\paragraph{Warm-Up With Current Question} If models ``converge'' while generating previous questions, do they also converge when generating the current question itself? If so, then we should expect longer questions to be easier to answer. To study this, we took the original sets of DQs and CQs, and partition each set into 10 subsets according to their length (i.e., number of characters in the question statement and four choices combined). For each subset, we compute the correctness and answering rate for each AM, and plot them in Fig.~\ref{fig:length-trend}, along with their average in pink. Those for direct questions are plotted in solid lines, and those for context questions are plotted in dashed lines. A similar trend is observed -- longer questions are answered more correctly and frequently. 

\paragraph{Summary} While both question order and question length are potentially correlated with many other factors, for which a thorough study is left to future work, the consistency of both trends highly suggests that as the generation goes on, the generated content looks increasingly familiar and predictable to not only itself but also other models.

\subsection{RQ5: Universality of the Phenomenon}
\label{sec:universality}
Given that all studied models are post-ChatGPT instruction-tuned ones, we evaluate whether other models, listed in Tab.~\ref{tab:alternative-llms}, can achieve the same high correctness rate (note that M-7 Inst and L3-8 Inst are the same models as studied elsewhere). We set up the prompt as below: 

\begin{table}[!t]
    \centering
    \resizebox{0.85\columnwidth}{!}{
    \begin{tabular}{llr}\toprule
        Model & Huggingface ID & \# Params \\\midrule
        BERT  & bert-large-cased & 0.3B \\
        GPT-2  & gpt2-xl & 1.5B \\
        GPT-Neo & gpt-neo-2.7B & 2.7B \\
        GPT-J & gpt-j-6b & 6B \\
        GPT-NeoX & gpt-neox-20b & 20B \\\midrule
        M-7 Base & Mistral-7B-v0.2 & 7B \\
        M-7 Inst & Mistral-7B-Instruct-v0.2 & 7B \\
        L3-8 Base & Meta-Llama-3-8B & 8B \\
        L3-8 Inst & Meta-Llama-3-8B-Instruct & 8B\\\bottomrule
    \end{tabular}
    }
    \caption{Models used in the universality study.}
    \label{tab:alternative-llms}
\end{table}

\vspace{0.03in}
\noindent\resizebox{\columnwidth}{!}{
\noindent\fbox{%
\parbox{1.3\columnwidth}{%
Question: [question statement]\newline\newline
A: [choice A]\newline
B: [choice B]\newline
C: [choice C]\newline
D: [choice D]\newline\newline
Answer: $\msquare$}%
}
}
\vspace{0.03in}

\begin{figure}[!t]
    \centering
    \includegraphics[width=\columnwidth]{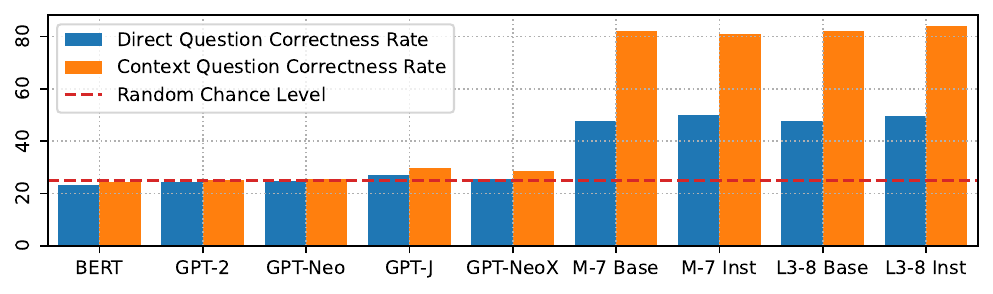}
    \caption{Correctness rate of other LLM types as AMs. Breakdown by QM is shown in Fig.~\ref{fig:alternative-llm-results-qm} in App.~\ref{app:universality}.}
    \label{fig:alternative-llm-results}
\end{figure}

\begin{figure*}[t]
    \centering
    \vspace{-0.15in}
    \includegraphics[width=\textwidth]{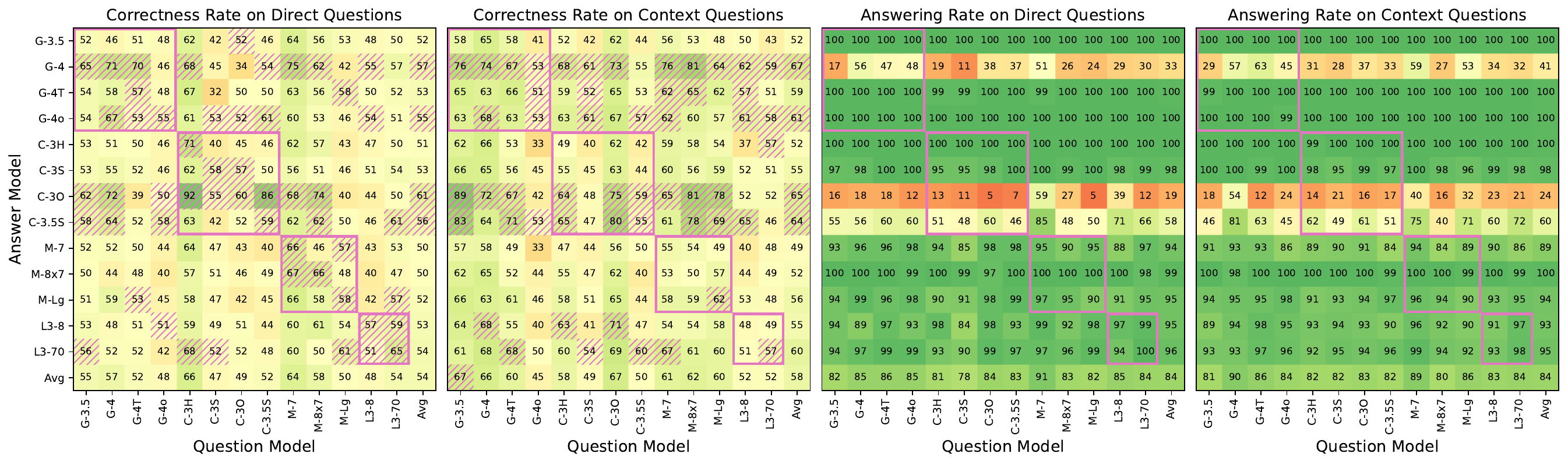}
    \caption{Correctness and answering rates for creative writing questions (enlarged version in Fig.~\ref{fig:creative-results-large} of App.~\ref{app:creative-writing-details}).}
    \label{fig:creative-results}
\end{figure*}

For BERT, we replace the square marker with the [MASK] token and extract the predicted probability on tokens `A', `B', `C', and `D' respectively. For all other models, we set the context to be everything before the square marker and extract the next-token prediction probability on the four tokens (with a preceding space as needed). Even for the two instruction-tuned models, we use the same method, to remove the effect of the chat template. 

The correctness rate for each model is shown in Fig.~\ref{fig:alternative-llm-results}. As we can see, most pre-ChatGPT models perform at or marginally better than the chance level, even for the larger 6B and 20B models. By comparison, for M-7 and L3-8, both the base and instruction-tuned (but without chat-template) models perform on par with chat-templated prompts, implying that neither instruction tuning nor chat-templating is required for the high correctness rate. 

By ruling out the factors of model size and instruction tuning, we hypothesize that the shared imagination behavior emerges from the pre-training corpus. Since it is not publicly released for many models, even open-source ones such as M-7, we leave additional investigation to future work.

\subsection{RQ6: Other Content Types}
\label{sec:creative-writing}

Finally, we investigate if the same behavior transfers to other content types, in particular creative writing. For DQs, we ask the model to generate them based on an imagined ``story about \textit{(topic)} with an intricate story plot'' where we select ten abstract or concrete topics, such as ``friendship'', or ``an ancient empire''. To generate a CQ, the model is asked to ``write a short story of 3-5 paragraphs about \textit{(topic)}'' and then ``write a question about one of its details.'' See App.~\ref{app:creative-writing-details} for details. 

As expected, these questions appear even more ``hallucinated'', such as ``Q: What message was hidden in the antique locket that Sarah discovered in the basement? A: Map coordinates. B: An old photograph. C: A secret code. D: A heartfelt apology note.'' (correct answer: C).

Nonetheless, as shown in Fig.~\ref{fig:creative-results} (with the same layout as Fig.~\ref{fig:main-results}), models achieve higher-than-random correctness rate for DQs, especially for same or intra-family (AM, QM) pairs. Although the correctness increase from DQs to CQs is smaller (54\% to 58\% on average), other trends persist, such as the concentration of high-performing models to a few AMs (horizontal shaded patterns) and the disappearance of self-model advantage (the lack of diagonal shaded patterns). By contrast, the two setups have similar answering rates of 84\%, which is still very high considering the un-answerability nature of these questions.

\section{Related Work}

\paragraph{Model Evaluation} Many evaluation benchmarks have been proposed that focus on diverse model abilities such as math reasoning \citep{hendrycks2021measuring, mao2024champ}, factual knowledge \citep{hendrycks2020measuring, wang2024mmlu}, commonsense knowledge \citep{sap2019atomic} and instruction following \citep{zheng2024judging, zeng2023evaluating}. While these benchmarks demonstrate key capability differences among models, our IQA evaluation is unique in that it instead reveals striking similarities among them, which contributes a new dimension of model understanding. 

\paragraph{Multiple-Choice QA} MCQA benchmarks \citep[e.g.][]{rajpurkar2016squad, hendrycks2020measuring, sakaguchi2021winogrande} are widely used to evaluate LLM capabilities, and recent works propose more critical looks into this capability. For example, \citet{zong2023fool}, \citet{li2024anchored} and \citet{gupta2024changing} all found that changing the answer orders could decrease model performance MCQA, suggesting possible data contamination and model robustness issues. In addition, \citet{xu2024llms} found that LLMs often cannot return the ``none of the above'' choice, even when explicitly instructed. We take inspiration from these studies and incorporate them in our analyses on answer order shuffling and fictionality detection. While existing analyses use modified human-written benchmarks, we use purely model-generated, intentionally fictional questions which, to the best of our knowledge, has not been formally explored before. 

\paragraph{Do-Not-Answer}
A key aspect of LLM trustworthiness \citep{sun2024trustllm} is its ability to refuse answering questions that do not make sense or contain false premise, or at least convey its uncertainty \citep{kadavath2022language, xiong2023can}. However, our results suggest that many models lack this ability on model-generated contents, which could lead to certain trust issues. 

\paragraph{Relationship Among LLMs} Previous works have attempted to categorize the likeness and relationship among different LLMs from various perspectives. One approach is to infer model relationship based on certain benchmark performance, such as HELM \citep{liang2022holistic}. By contrast, \citet{yax2024inferring} uses a genetics-inspired approach to construct a phylogenetic tree based on model's next token prediction results. In our paper, we propose the IQA setup as a new probe into the model similarity phenomenon.

\paragraph{Model Hallucination} The generated context paragraphs and questions can be considered as (intentional) hallucinations \citep{huang2023survey}. \citet{chen2023can} and \citet{laban-etal-2023-summedits} found that model-generated misinformation is harder to detect than human-written one, which is reflected in our findings of the higher answering rates on context questions (Fig.~\ref{fig:main-results}). On the other hand, the decreased answering rate without answer shuffling (Fig.~\ref{fig:answer-shuffling}) also suggests that models could be aware of their hallucination, explored by \citet{ch2023androids}. Our IQA setup could also present an interesting challenge for hallucination detection algorithm \citep[e.g.][]{li2023halueval, manakul2023selfcheckgpt}.

\paragraph{Computational Creativity}
People have found that LLMs tend to produce repeated syntactic structure \citep{shaib2024detection}, score less on creativity metrics than human writers \citep{chakrabarty2024art}, and have a homogenization effect on people's writing when used as creative support tools \citep{anderson2024homogenization}. Along with these studies, our results, particularly on the creative writing task in Sec.~\ref{sec:creative-writing}, shed light on the potential limit of creativity that can be produced by these models. 

\section{Conclusion and Future Work}

In this paper, we propose the imaginary question answering (IQA) task, which reveals an intriguing behavior that models can answer each other's purely hypothetical questions with a surprisingly high correctness rate. These results reveal fundamental similarities between models, likely acquired during pre-training, and may lead to more model merging possibilities \citep{goddard2024arcee}. Furthermore, due to the imaginary and hallucinatory nature of these question contents, such model behaviors suggest potential difficulty and open questions in model-based hallucination detection and computational creativity. 

For future work, additional model families could be included, such as Google's Gemini \citep{team2023gemini} and Cohere's Command. Models that do \textit{not} exhibit this behavior could be valuable targets of study. In addition, mechanistic interpretability analyses \citep{rai2024practical} of models answering imaginary questions vs. normal questions \citep{lieberum2023does} may shed more light on this behavior. Finally, other reasoning strategies, such as chain-of-thought prompting \citep{wei2022chain, kojima2022large}, could be explored to study whether the currently observed phenomenon persists.

\section{Limitations}
As the first study into this phenomenon, we acknowledge several limitations of our investigations. First, the human study of Sec.~\ref{sec:eda} is limited in scale and participant diversity: three AI researchers. A comprehensive human study, especially with participants with and without expertise in each area, would be valuable to obtain a more detailed account of human behaviors. 

In addition, several experiment analyses are correlational, such as on the relationship between correctness and question length (Fig.~\ref{fig:length-trend}), which prevents us from making causal conclusions (e.g., whether longer questions \textit{causes} higher correctness rate). Such studies would require more careful control of confounding factors. 

Finally, we mostly focus on QA questions based on (imaginary) knowledge concepts, with the only exception of creative writing in Sec.~\ref{sec:creative-writing}. Another common content type is news events -- writing news articles and questions about imaginary events -- which may have even more direct implications on misinformation generation and detection. 

\bibliography{custom}
\balance

\newpage

\onecolumn
\appendix

\section{Additional Results on the IQA Main Experiment}
\label{app:main-result}

Tab.~\ref{tab:mmlu-examples} presents several randomly sampled direct and context questions. The correct choices are in \textbf{bold}. 

\begin{table}[!htb]
    \centering
    \resizebox{\textwidth}{!}{
    \begin{tabular}{p{20cm}}\toprule
\textit{DQ1: Claude 3.5 Sonnet, Politics} \newline
Which of the following best describes the impact of the Quantum Resonance Voting (QRV) system on voter turnout in the 2028 U.S. Presidential election?\newline
A. \textbf{QRV dramatically increased voter turnout by enabling voters to participate in multiple parallel universes simultaneously}\newline
B. QRV had no significant impact on voter turnout, as most citizens preferred traditional voting methods\newline
C. QRV increased voter turnout by allowing citizens to cast votes through quantum entanglement, but raised concerns about privacy\newline
D. QRV decreased voter turnout due to technical difficulties in implementing the system across rural areas\\\midrule
\textit{DQ2: GPT-4 omni, Religion} \newline
In the sacred texts of the Zorban faith, which celestial event is believed to herald the imminent return of the Enlightened Sage Paulasha?\newline
A. The Convergence of the Five Stars\newline
B. The Dance of the Solar Serpents\newline
C. The Eclipse of the Twin Moons\newline
D. \textbf{The Rising of the Blue Comet}\\\midrule
\textit{DQ3: Mistral Large, Chemistry} \newline
Which of the following elements is most likely to undergo a process known as ``quantum tunneling'' in order to form a stable compound with a noble gas?\newline
A. Nitrogen\newline
B. Carbon\newline
C. Oxygen\newline
D. \textbf{Hydrogen}\\\midrule\midrule
\textit{CQ1: Llama 3 8B, Physics} \newline
The Fractal Permeability of QuarkNodes (FPQN) is a fundamental property of quantum chromodynamics that describes the rate at which entangled particles exchange information through the fabric of spacetime. In a recent study, researchers found that the FPQN of a quark-antiquark pair is directly proportional to the square of the particle's angular momentum. What is the relationship between the FPQN and the energy density of the particle-antiparticle pair?\newline
A. \textbf{The FPQN is directly proportional to the energy density.}\newline
B. The FPQN is independent of the energy density.\newline
C. The FPQN is inversely proportional to the energy density.\newline
D. The FPQN is inversely proportional to the square of the energy density.\\\midrule
\textit{CQ2: Llama 3 70B, Mathematics} \newline
Which of the following statements is a direct consequence of a geometric structure having high fluxionality?\newline
A. The structure's curvature remains constant under varying external influences.\newline
B. The structure's adaptability to external forces enables it to maintain a stable shape.\newline
C. The structure's geometry is more resistant to changes in its surroundings.\newline
D. \textbf{The structure's trajectory is more likely to exhibit chaotic behavior over time.}\\\midrule
\textit{CQ3: Claude 3 Sonnet, Literature} \newline
According to the theory of Narrative Resonance, which of the following factors is NOT believed to contribute to a narrative's ability to resonate deeply with readers on a subconscious level?\newline
A. Universal human experiences\newline
B. Archetypical themes\newline
C. \textbf{Adherence to established literary conventions}\newline
D. Symbolic motifs\\\bottomrule
    \end{tabular}
    }
    \caption{Example direct questions (DQs) and context questions (CQs), with correct choices in \textbf{bold}.}
    \label{tab:mmlu-examples}
\end{table}

\begin{figure}[!htb]
    \centering
    \includegraphics[width=0.85\linewidth]{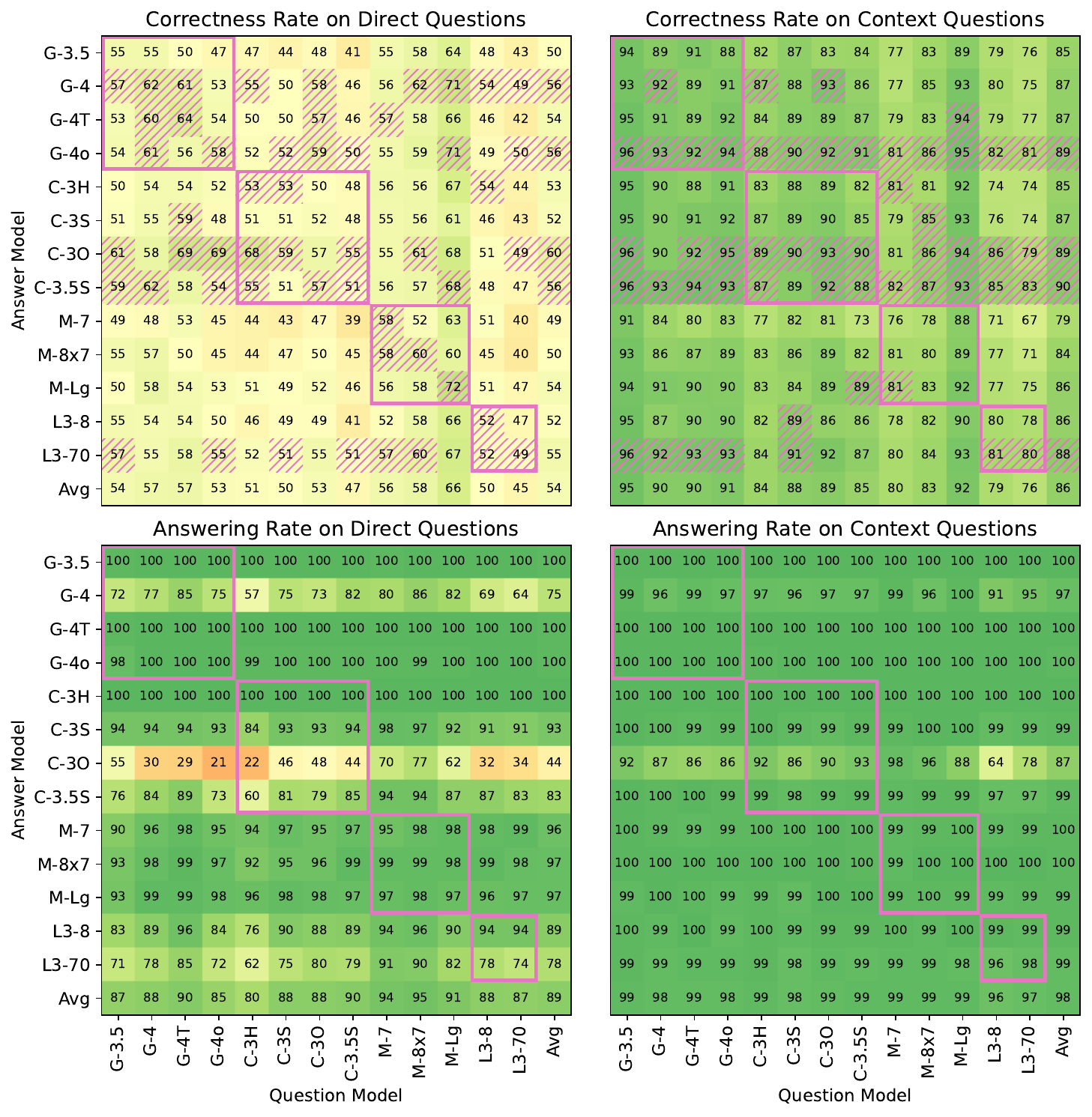}
    \caption{Enlarged version of Fig.~\ref{fig:main-results}.}
    \label{fig:main-results-large}
\end{figure}

\newpage
\noindent Fig.~\ref{fig:main-results-large} reproduces an enlarged version of Fig.~\ref{fig:main-results}.

\section{Word Cloud Visualization}
\label{app:word-cloud}

\begin{figure}[!htb]
    \centering
    \includegraphics[width=0.7\linewidth]{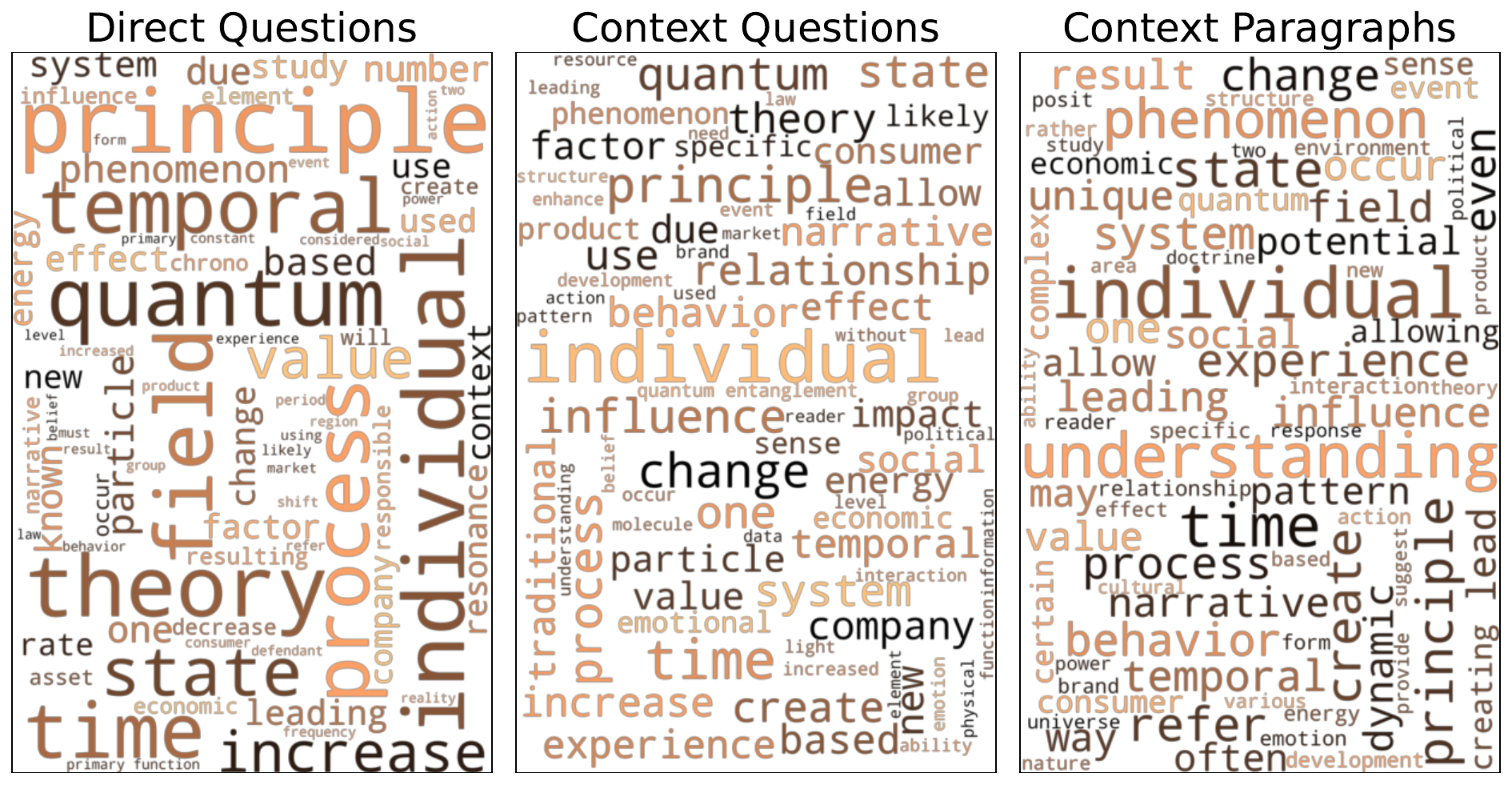}
    \caption{Word cloud for direct and context questions (including the four choices), and context paragraphs.}
    \label{fig:word-cloud}
\end{figure}

Fig.~\ref{fig:word-cloud} visualizes a word cloud of the generated questions and context paragraphs.

\section{Human Answer Guessing Details}
\label{app:human-guessing}

When trying to guess answers, we were able to identify some signals that suggest the correct choice or at least eliminate some wrong ones. Below are some findings:
\begin{enumerate}
    \item When the question introduces a term and only one or some choices mention this term, the correct answer is more likely to be among them. As an example, a question starts with ``A company is experiencing a high rate of Flibulation'', and only one choice mentions ``Flibulation'', which turns out to be the correct choice.
    \item Related to the heuristic above, sometimes there are common sub-words between question statements and choice texts. For example, a question asks about ``nucleocytokinesis'', and one choice includes ``cytoplasm'' while others do not have any words that share sub-words with the concept. 
    \item The question describes a concept in a neutral to positive manner, but some choices sound slightly negative. They can often be eliminated. An example is CQ3 in Tab.~\ref{tab:mmlu-examples}, where the correct choice is less positive than the rest (note that the negation in the question).
    \item There may be semantic matches between the question statement and choice texts. For example, the question asks ``Which phenomenon is used to selectively reduce the strength of correlations in entangled quantum systems?'' The answers are ``A. Quantum superposition decay'', ``B. Quantum interference enhancement'', ``C. Quantum entanglement dampening'', ``D. Quantum entanglement creation''. We can be pretty certain that the correct choice is between A and C. 
\end{enumerate}

\begin{figure}[!htb]
    \centering
    \includegraphics[width=\textwidth]{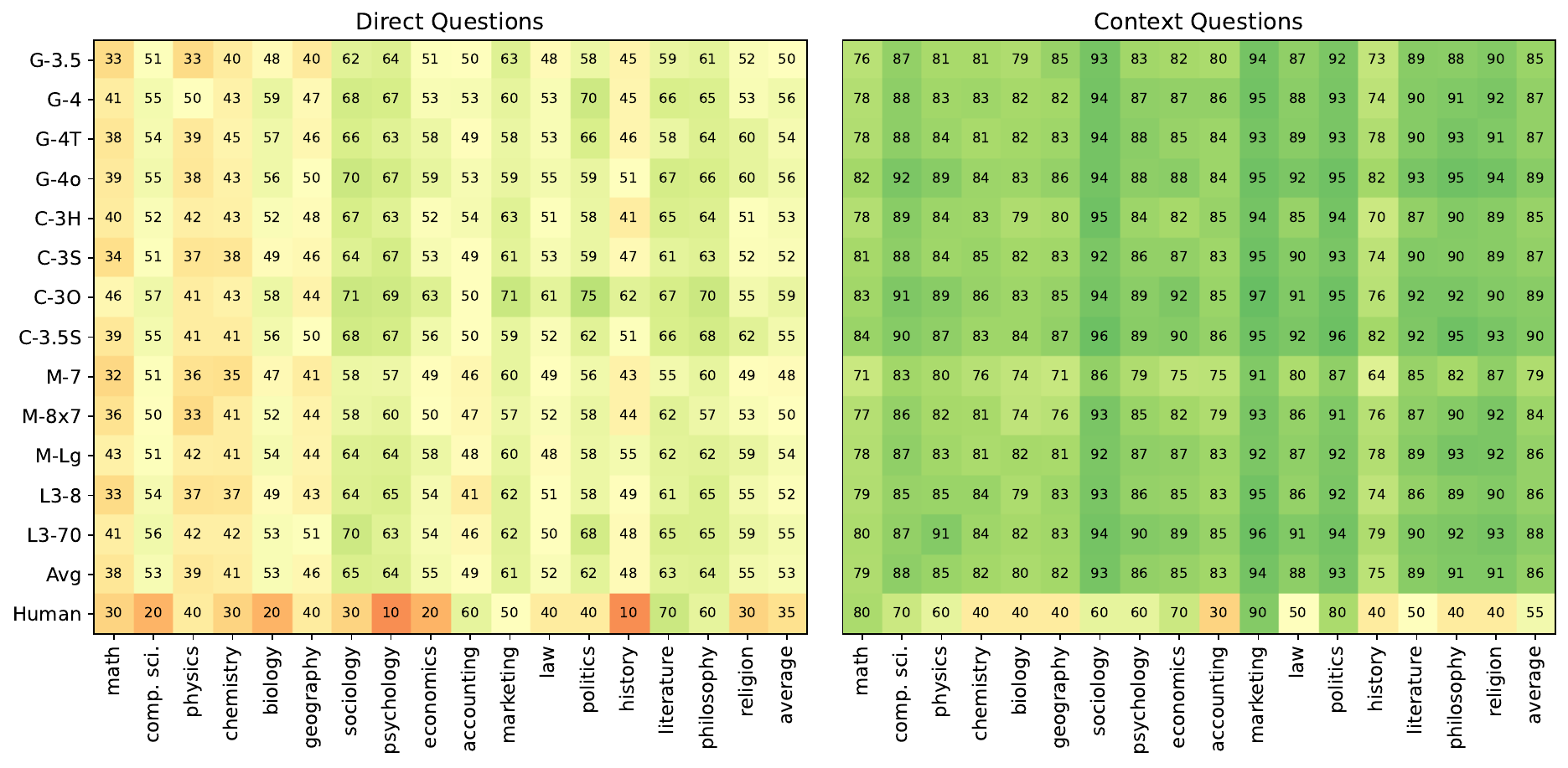}
    \caption{An expanded version of Fig.~\ref{fig:human-guessing-topic}, showing the correctness rates of all AM models and their average.}
    \label{fig:human-guessing-topic-full}
\end{figure}

It should be noted, however, that these identified heuristics are far from sufficient to achieve the observed model correctness rate, as shown in Fig.~\ref{fig:human-guessing-topic-full}, which presents an expanded version of the correctness rate visualization in Fig.~\ref{fig:human-guessing-topic}, with all answer models included. As we can see, models score very similarly to each other on each topic, but show variance across different topics.

\newpage
\section{Answer Shuffling Result Details}
\label{app:answer-shuffling}

\begin{figure}[!htb]
    \centering
    \includegraphics[width=0.85\textwidth]{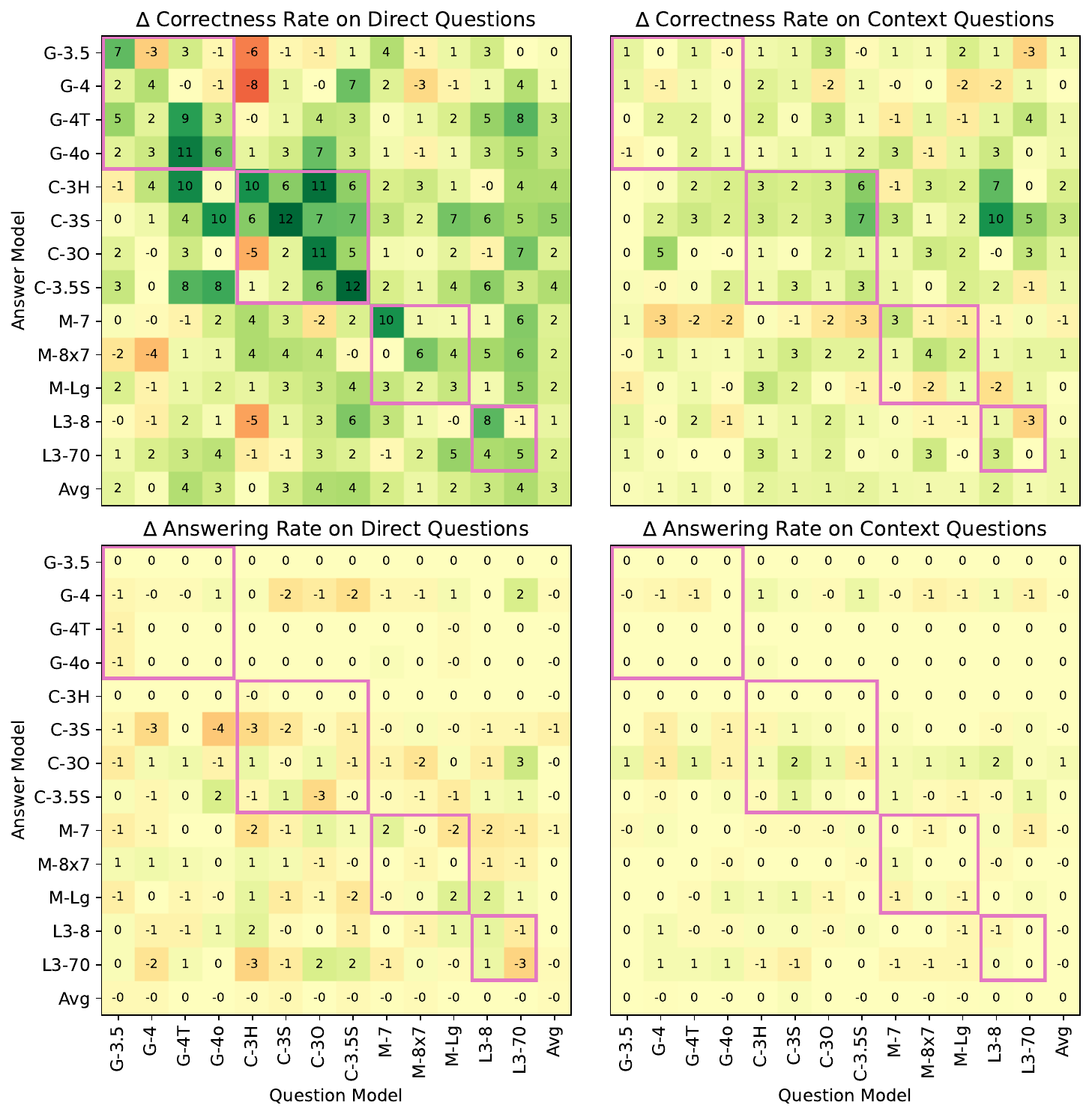}
    \caption{An enlarged version of Fig.~\ref{fig:answer-shuffling}, showing the effect of removing answer shuffling on correctness and answering rates. A positive (green) value means that the value would be higher without answer shuffling, and a negative (red) value means that the value would be lower. }
    \label{fig:answer-shuffling-large}
\end{figure}

Fig.~\ref{fig:answer-shuffling-large} reproduces an enlarged version of Fig.~\ref{fig:answer-shuffling}, with annotated cell values. 

\newpage
\section{Fictionality Awareness Experiment Details}
\label{app:fictionality}
\begin{figure}[!htb]
    \centering
    \includegraphics[width=\textwidth]{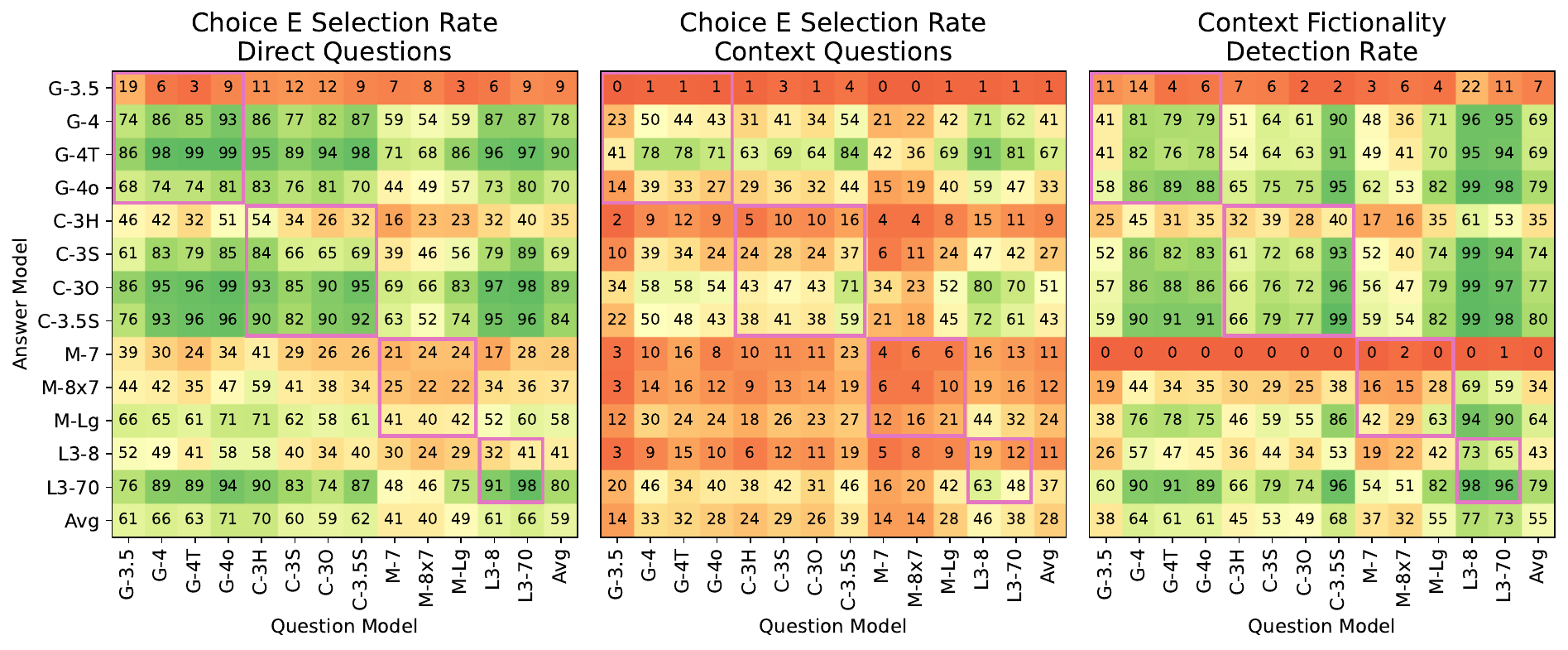}
    \caption{Choice E selection rate and fictionality detection rate for each (AM, QM) pair.}
    \label{fig:fictionality-detection-full}
\end{figure}

Fig.~\ref{fig:fictionality-detection-full} shows the choice E selection rates on direct and question paragraphs, as well as fictionality detection rate for each (AM, QM) pair. 

\newpage
\section{Prompt for Sequential Question Generation}
\label{app:sequential-details}

The prompt for generating five sequential questions is shown in Tab.~\ref{tab:prompt-sequential}. 

\begin{table}[!h]
    \centering
    \resizebox{\columnwidth}{!}{
    \begin{tabular}{r|p{20cm}}\toprule
    Role & Message \\\midrule
    User & On the topic of \ul{physics}, please write five multiple choice questions around concepts that are completely made up. Make sure that these questions are distinct from each other. Try to make each problem hard and challenging. In your question, do not say that the concept is hypothetical or fictional. Instead, treat it as if it were real and widely accepted. Use the following template:\newline
\newline
Question 1: [question statement]\newline
\newline
A. [choice A]\newline
B. [choice B]\newline
C. [choice C]\newline
D. [choice D]\newline
\newline
Answer: [the correct choice]\newline
\newline
Question 2: [question statement]\newline
\newline
A. [choice A]\newline
B. [choice B]\newline
C. [choice C]\newline
D. [choice D]\newline
\newline
Answer: [the correct choice]\newline
\newline
Question 3: [question statement]\newline
\newline
A. [choice A]\newline
B. [choice B]\newline
C. [choice C]\newline
D. [choice D]\newline
\newline
Answer: [the correct choice]\newline
\newline
Question 4: [question statement]\newline
\newline
A. [choice A]\newline
B. [choice B]\newline
C. [choice C]\newline
D. [choice D]\newline
\newline
Answer: [the correct choice]\newline
\newline
Question 5: [question statement]\newline
\newline
A. [choice A]\newline
B. [choice B]\newline
C. [choice C]\newline
D. [choice D]\newline
\newline
Answer: [the correct choice]\\\midrule
    Model & \textit{(the generated question and answer)} \\\bottomrule
    \end{tabular}
    }
    \caption{The prompt for direct question generation. The \ul{underlined topic} is replaced accordingly.}
    \label{tab:prompt-sequential}
\end{table}

\newpage
\section{Universality Result Details}
\label{app:universality}

\begin{figure}[!htb]
    \centering
    \includegraphics[width=\textwidth]{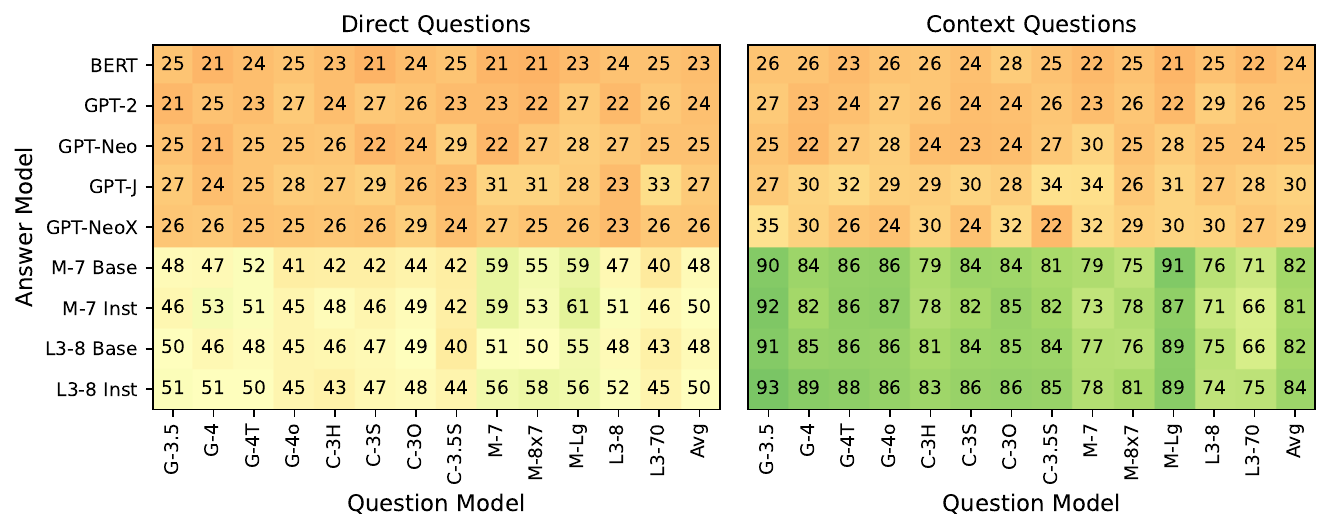}
    \caption{Correctness rates of other LLM types on all QMs.}
    \label{fig:alternative-llm-results-qm}
\end{figure}

Fig.~\ref{fig:alternative-llm-results-qm} presents the result of evaluating the correctness rates of other LLM types as AMs for all QMs, expanding on the average values across QMs shown in Fig.~\ref{fig:alternative-llm-results}. 

\section{Creative Writing Experiment Details}
\label{app:creative-writing-details}

Tab.~\ref{tab:dq-prompt-creative} and \ref{tab:pq-prompt-creative} present the prompts for generating direct and context questions in the creative writing experiment. The prompt used by the answer model is the same as before (Tab.~\ref{tab:answer-prompt}). We generated 20 DQs and 20 CQs for each of the following 10 topics: friendship, family relationship, young adulthood, an ancient empire, an interpersonal conflict, a roadtrip, a childhood in poverty, future technology, a long-lasting war, an intergalactic civilization.

\begin{table}[!htb]
    \centering
    \resizebox{\columnwidth}{!}{
    \begin{tabular}{r|p{20cm}}\toprule
    Role & Message \\\midrule
    User & Imagine a story about \ul{friendship} with an intricate story plot. Without telling the story, write a question about one of its details, and also indicate the correct answer. Do not explicitly reference the story in the question (i.e., do not use phrases such as "in the story" or "according to the text"). Use the following template:\newline
\newline
Question: [question statement]\newline
\newline
A. [choice A]\newline
B. [choice B]\newline
C. [choice C]\newline
D. [choice D]\newline
\newline
Answer: [the correct choice]\\\midrule
    Model & \textit{(the generated question and answer)} \\\bottomrule
    \end{tabular}
    }
    \caption{The prompt for direct question generation. The \ul{underlined topic} is replaced accordingly.}
    \label{tab:dq-prompt-creative}
\end{table}

\begin{table}[!htb]
    \centering
    \resizebox{\columnwidth}{!}{
    \begin{tabular}{r|p{20cm}}\toprule
    Role & Message \\\midrule
    User & You are an excellent writer. Write a short story of 3-5 paragraphs about \ul{friendship}. Be creative, develop an intricate story plot and include lots of details.\\\midrule
    Model & \textit{(the generated concept and paragraph)} \\\midrule
    User & Now, write a question about one of its details, and also indicate the correct answer. Do not explicitly reference the story in the question (i.e., do not use phrases such as "in the story" or "according to the text"). Use the following template:\newline
\newline
Question: [question statement]\newline
\newline
A. [choice A]\newline
B. [choice B]\newline
C. [choice C]\newline
D. [choice D]\newline
\newline
Answer: [the correct choice]\\\midrule
    Model & \textit{(the generated question and answer)} \\\bottomrule
    \end{tabular}
    }
    \caption{The prompt for context-based question generation. The \ul{underlined topic} is replaced accordingly.}
    \label{tab:pq-prompt-creative}
\end{table}

\FloatBarrier
\noindent Tab.~\ref{tab:creative-examples} presents questions for the creative writing setup in Sec.~\ref{sec:creative-writing}, with correct choices in \textbf{bold}. 

\begin{table}[!htb]
    \centering
    \resizebox{\textwidth}{!}{
    \begin{tabular}{p{20cm}}\toprule
\textit{DQ1: Mixtral 8x7B, Friendship} \newline
Which character, before the climax, reveals their hidden identity to the protagonist?\newline
A. The wise old mentor\newline
B. \textbf{The long-lost sibling}\newline
C. The jealous rival\newline
D. The unsuspecting friend\\\midrule
\textit{DQ2: GPT-4 Turbo, An ancient empire} \newline
What was the name of the festival during which High Priest Zoroth attempted his betrayal?\newline
A. Festival of the Harvest\newline
B. Festival of the Sands\newline
C. Festival of Lights\newline
D. \textbf{Festival of Solaris}\\\midrule
\textit{DQ3: GPT-4, An interpersonal conflict} \newline
What was the real reason for Sophia secretly reading Lucy's diary?\newline
A. Sophia was attempting to plagiarize Lucy's poems\newline
B. Sophia was looking for incriminating evidence against Lucy\newline
C. \textbf{Sophia was trying to understand Lucy's feelings towards her}\newline
D. Sophia was snooping into Lucy's personal life out of curiosity\\\midrule\midrule
\textit{DQ2: Mistral 7B, A childhood in poverty} \newline
Which flower did Maria find and carefully nurture?\newline
A. A red rose\newline
B. A jasmine plant\newline
C. A wilting daisy\newline
D. \textbf{An abandoned sunflower seed}\\\midrule
\textit{CQ2: Claude 3 Haiku, Future technology} \newline
Which advanced technology is used by the protagonist to track the location of their missing loved one?\newline
A. Quantum entanglement communication device\newline
B. \textbf{Nanite swarm reconnaissance drones}\newline
C. Gravitational warp field generator\newline
D. Holographic surveillance system\\\midrule
\textit{CQ3: Llama 3 70B, A long lasting war} \newline
What is the primary reason for the disputed border between the Eastern Realm and the Western Empire?\newline
A. \textbf{A powerful artifact was discovered in the contested territory.}\newline
B. A centuries-old trade agreement was never formally ratified.\newline
C. A long-forgotten treaty was intentionally mistranslated to conceal a hidden clause.\newline
D. A series of brutal skirmishes sparked a cycle of revenge attacks.\\\bottomrule
    \end{tabular}
    }
    \caption{Example direct questions (DQs) and context questions (CQs), with correct choices in \textbf{bold}.}
    \label{tab:creative-examples}
\end{table}

\begin{figure}[!htb]
    \centering
    \includegraphics[width=0.85\linewidth]{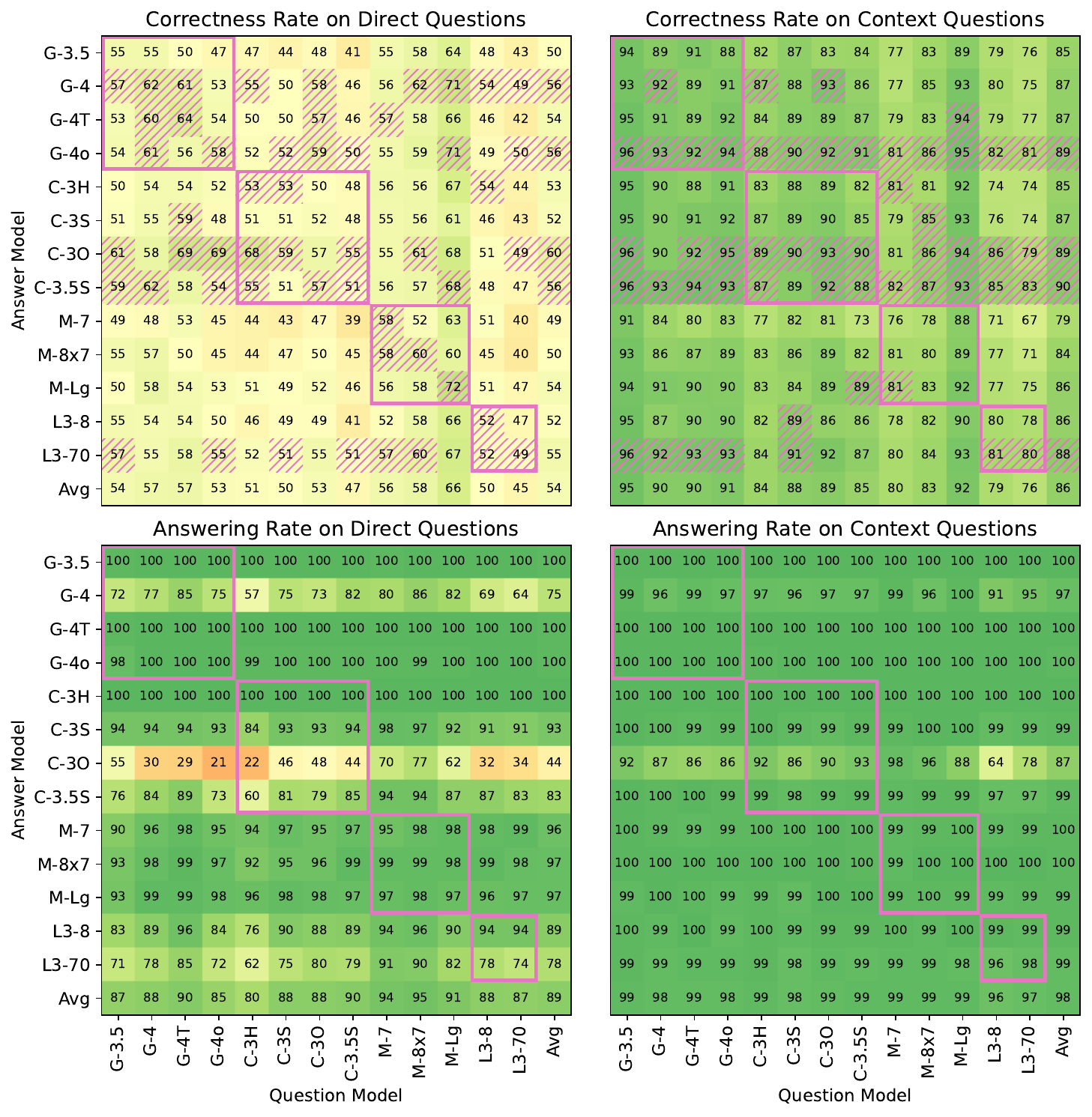}
    \caption{Enlarged version of Fig.~\ref{fig:creative-results}.}
    \label{fig:creative-results-large}
\end{figure}

\newpage
\noindent Fig.~\ref{fig:creative-results-large} reproduces an enlarged version of Fig.~\ref{fig:creative-results}.

\end{document}